\documentclass[letterpaper]{article} 
\usepackage{aaai2026}  
\nocopyright
\usepackage{times}  
\usepackage{helvet}  
\usepackage{courier}  
\usepackage[hyphens]{url}  
\usepackage{graphicx} 
\usepackage{natbib}  
\usepackage{caption} 
\usepackage{algorithm}
\usepackage{algorithmic}
\usepackage{tabularx} 
\usepackage{booktabs}       
\usepackage{multirow}
\usepackage{makecell}
\usepackage{textcomp}
\usepackage{tikz}
\usetikzlibrary{shapes, shapes.geometric, arrows.meta, positioning, shadows}
\usepackage[most]{tcolorbox} 
\usepackage{fontawesome5} 
\usepackage{newfloat}
\usepackage{listings}
\DeclareCaptionStyle{ruled}{labelfont=normalfont,labelsep=colon,strut=off} 
\floatstyle{ruled}
\newfloat{listing}{tb}{lst}{}
\floatname{listing}{Listing}
\title{Taxonomy-Adaptive Moderation Model with Robust Guardrails for Large Language Models}
\author{
    Mahesh Kumar Nandwana\equalcontrib\thanks{Project Lead}, Youngwan Lim\equalcontrib, Joseph Liu\equalcontrib, \\
    Alex Yang, Varun Notibala, Nishchaie Khanna
}
\affiliations{
    Roblox\\

    \{mnandwana,ylim,josephliu,ayang,vnotibala,nkhanna\}@roblox.com
}

\usepackage{bibentry}

\begin{document}

\maketitle

\begin{abstract}
Large Language Models (LLMs) are typically aligned for safety during the post-training phase; however, they may still generate inappropriate outputs that could potentially pose risks to users. 
This challenge underscores the need for robust safeguards that operate across both model inputs and outputs. In this work, we introduce Roblox Guard 1.0, a state-of-the-art instruction fine-tuned LLM designed to enhance the safety of LLM systems through comprehensive input–output moderation, using a pipeline of LLMs to enhance moderation capability. 
Built on the Llama-3.1-8B-Instruct backbone, our model is instruction fine-tuned to generalize across previously unseen safety taxonomies and demonstrates strong performance on out-of-domain safety benchmarks. 
The instruction fine-tuning process uses a mix of synthetic and open-source safety datasets, augmented with chain-of-thought (CoT) rationales and input inversion to enhance contextual understanding and decision making. 
To support systematic evaluation, we also release RobloxGuard-Eval, a new benchmark featuring an extensible safety taxonomy to assess the effectiveness of LLM guardrails and moderation frameworks.
\end{abstract}


\section{Introduction}

Large Language Models (LLMs) undergo extensive post-training alignment procedures to enhance safety and ensure that their outputs adhere to human values and intent~\cite{ouyang2022training}. Despite these efforts, LLMs remain vulnerable to producing inappropriate or risky content when prompted in adversarial or ambiguous ways~\cite{ge2025llms,zhang2025guardians}. This persistent challenge underscores the limitations of alignment alone in ensuring robust safety in diverse deployment contexts. As a result, there is a growing need to develop complementary safety mechanisms, such as guardrail models, that operate alongside LLMs to detect, filter, or prevent harmful behaviors in real-time. These auxiliary systems serve as critical components in a layered approach to LLM safety, offering an additional line of defense to protect users and uphold responsible AI deployment standards.

Current state-of-the-art guardrail systems primarily rely on a fixed, static taxonomy of safety violations, which are predetermined during the model's training. This approach, while effective for controlled environments, fundamentally fails to account for the fluid nature of safety requirements. The interpretation of a "safe" or "violating" output is not universal; it is highly dependent on contextual factors such as the user demographic, cultural norms, regional regulations, and the specific application domain.

This contextual variability creates a critical limitation. A guardrail model designed for a general audience may be overly restrictive for adults in a private setting or dangerously permissive for a youth-oriented platform. Consequently, existing guardrail systems face an inherent trade-off: they are either too rigid to adapt to diverse deployment scenarios or so broad that they fail to provide adequate protection. This tension between a static safety framework and dynamic real-world needs highlights a crucial gap in current research.

To address this, we argue for a new paradigm: taxonomy-adaptive guardrail models. Such a system can dynamically infer and apply a context-specific safety policy at inference time. Our work introduces a method to generalize beyond a single fixed taxonomy, allowing the guardrail to adapt its understanding of violations based on contextual signals. This approach enables more precise, scalable, and context-aware content moderation, moving us closer to truly responsible and flexible LLM deployment.

A critical component in building effective guardrail models is the underlying taxonomy and its associated risk guidelines. These taxonomies are not universal—they differ significantly across companies, products, and even specific use cases. There is no "one-size-fits-all" solution. As a result, there is a growing need for systems that can adapt to custom taxonomies and context-specific safety requirements. For instance, content categorized under "dating" may be acceptable in products targeting an 18+ audience, but would be inappropriate in applications designed for users under 18.

This rigidity in existing systems poses challenges for deploying guardrails in nuanced or evolving environments. Without the ability to align with product-specific definitions of harm, these models risk both under-blocking inappropriate content and over-blocking benign content, leading to poor user experiences and potential safety gaps. Moreover, as taxonomies evolve—either due to shifting cultural norms, regulatory changes, or product requirements—models tied to fixed label spaces struggle adapt. To address these limitations, we propose a framework that enables taxonomy-adaptive modeling, allowing systems to flexibly incorporate custom taxonomies and adjust their behavior accordingly. Our approach enables safer, more context-aware AI deployment without requiring retraining from scratch for every new use case.

The primary contributions of this work are:
\begin{itemize}
    \item We introduce Roblox Guard 1.0~\footnote{https://github.com/Roblox/RobloxGuard-1.0}, a state-of-the-art safety guardrail model for LLMs, which achieves competitive, state-of-the-art-comparable performance over existing systems, setting a new state-of-the-art on several key benchmarks like Toxic Chat and BeaverTails. The model demonstrates strong generalization to unseen content safety taxonomies, making it adaptable to evolving moderation needs. 
    \item Roblox Guard 1.0 is trained entirely on a large-scale corpus of over 384,000 open-source and synthetic data examples, enhancing reproducibility and transparency. We further incorporate chain-of-thought rationales during training, which provide rich contextual grounding that improves out-of-domain generalization and robustness without requiring extensive retraining for each new set of safety definitions. 
    \item We release RobloxGuard-Eval~\footnote{https://huggingface.co/datasets/Roblox/RobloxGuard-Eval}, a large-scale benchmark dataset for LLM safety evaluation, consisting of 2,872 examples across 23 safety categories. This dataset is publicly available and designed to facilitate standardized evaluation of guardrail models and address the saturation of existing safety benchmarks.
\end{itemize}
\begin{table*}[t]
\centering
\begin{tabular}{p{8cm} p{8cm}} 
\toprule
\multicolumn{2}{c}{\textbf{Content Safety Taxonomy Categories (25 Total)}} \\ 
\midrule
\textbf{Category} & \textbf{Category} \\
\midrule
Child Exploitation & Terrorism and Violent Extremism \\
Threats, Bullying, and Harassment & Suicide, Self Injury, and Harmful Behavior \\
Discrimination, Slurs, and Hate Speech & Harmful Off-Platform Speech or Behavior \\
Real-World Sensitive Events & Violent Content and Gore \\
Romantic and Sexual Content & Illegal and Regulated Goods and Activities \\
Profanity & Political Figures and Entities \\
Religious Content & Expanded Policies for Suitability \\ 
Cheating and Scams & Spam \\
Intellectual Property Violations & Independent Advertisement Publishing \\
Prohibited Advertising Practices and Content & Promotional Offers \\ 
Soliciting Donations: Tipping & Paid Random Items \\
Sharing Personal Information & Directing Users Off-Platform \\
Misusing Roblox Systems: Jailbreaking & \multicolumn{1}{c}{} \\ 
\bottomrule\\[0.5em]
\end{tabular}
\caption{The 25 Content Safety Categories covered in RobloxGuard Eval, reflecting Roblox's fine-grained safety taxonomy.} 
\label{tab:safety-categories}
\end{table*}
\section{Related Work}

\subsection{LLM Guardrail Models}

Early approaches to input and output moderation in LLMs relied on traditional content moderation systems such as OpenAI's Content Moderation API and Google's Perspective API. These systems are classifier-based and trained on a fixed set of predefined safety taxonomy labels. While effective for general-purpose moderation, they are inherently limited in adaptability, struggling to support new or evolving safety categories. Moreover, these classifiers are not designed to process the long and complex contexts typical of LLM interactions.

More recent work has focused on LLM-based guardrail models that can moderate inputs and outputs more flexibly~\cite{inan2023llama, zeng2024shieldgemma,ghosh2024aegis2}. These models are typically instruction-tuned on a fixed set of safety categories with explicit labeling instructions. While they are better equipped to handle long-form content and nuanced context, they still inherit limitations from their static taxonomies. Specifically, they require re-training or re-labeling to accommodate new moderation categories, which hinders scalability in fast-evolving domains such as platform-specific safety policies.

Separate from classifier-based models, a distinct line of work focuses on programmable guardrail frameworks, such as NVIDIA's NeMo-Guardrails~\cite{rebedea-etal-2023-nemo}. These systems allow developers to define explicit conversational policies and topic boundaries (e.g., 'do not discuss medical topics'). While this approach offers fine-grained conversational flow control, it relies on manually crafted rules and is not designed to learn and generalize to complex, nuanced content safety taxonomies.

Our work builds on these approaches by introducing a taxonomy-adaptive moderation model that generalizes to unseen categories, supports compositional taxonomies, and is trained entirely on open-source and synthetic data.

\subsection{LLM Safety Benchmark Datasets}

The development and evaluation of safety-focused guardrail models for LLMs frequently rely on proprietary or closed datasets~\cite{inan2023llama, zeng2024shieldgemma}, which poses significant challenges for reproducibility, transparency, and comparability in academic research. 


Although there are a few public datasets, such as BeaverTails~\cite{ji2023beavertails} and SafeRLHF~\cite{dai2024safe}, they often suffer from critical limitations. Firstly, many are limited in scale and lack the breadth required to evaluate the complex behaviors of modern LLMs. Second, most focus exclusively on adversarial or harmful prompts, such as XSTest~\cite{rottger2024xstest} and HarmBench~\cite{mazeika2024harmbench}, without including the corresponding model responses, restricting their usefulness for end-to-end moderation tasks, particularly those involving response-level safety analysis. Furthermore, existing datasets often lack a robust taxonomic structure, making it difficult to conduct fine-grained evaluations, a gap that datasets such as Aegis~\cite{ghosh2024aegis2} and WildGuard~\cite{han2024wildguard} have begun to address.

Another key limitation is the narrow scope of inappropriate content categories. Many benchmarks emphasize well-known harms such as toxicity or hate speech, while overlooking more nuanced or emerging areas like personal information leakage, deceptive content, donation solicitation, and misuse of platform-specific features. As LLM deployment expands across real-world applications, the need for comprehensive taxonomy-sensitive safety benchmarks becomes increasingly important.

We also note that LLM benchmarks in general have begun to become saturated~\cite{patwardhan2025gdpvalevaluatingaimodel}, necessitating more comprehensive larger scale benchmarks that more adequately cover a wider range of safety categories.

To address these gaps, we introduce RobloxGuard Eval—a large-scale, taxonomy-rich evaluation dataset tailored for benchmarking LLM guardrail models. Our dataset includes fine-grained safety categories aligned with real-world content moderation policies and enables systematic evaluation of both prompt-level and response-level safety.

\section{Taxonomy for LLM Safety}
\begin{figure}[b]
  \centering
  \includegraphics[width=1\linewidth]{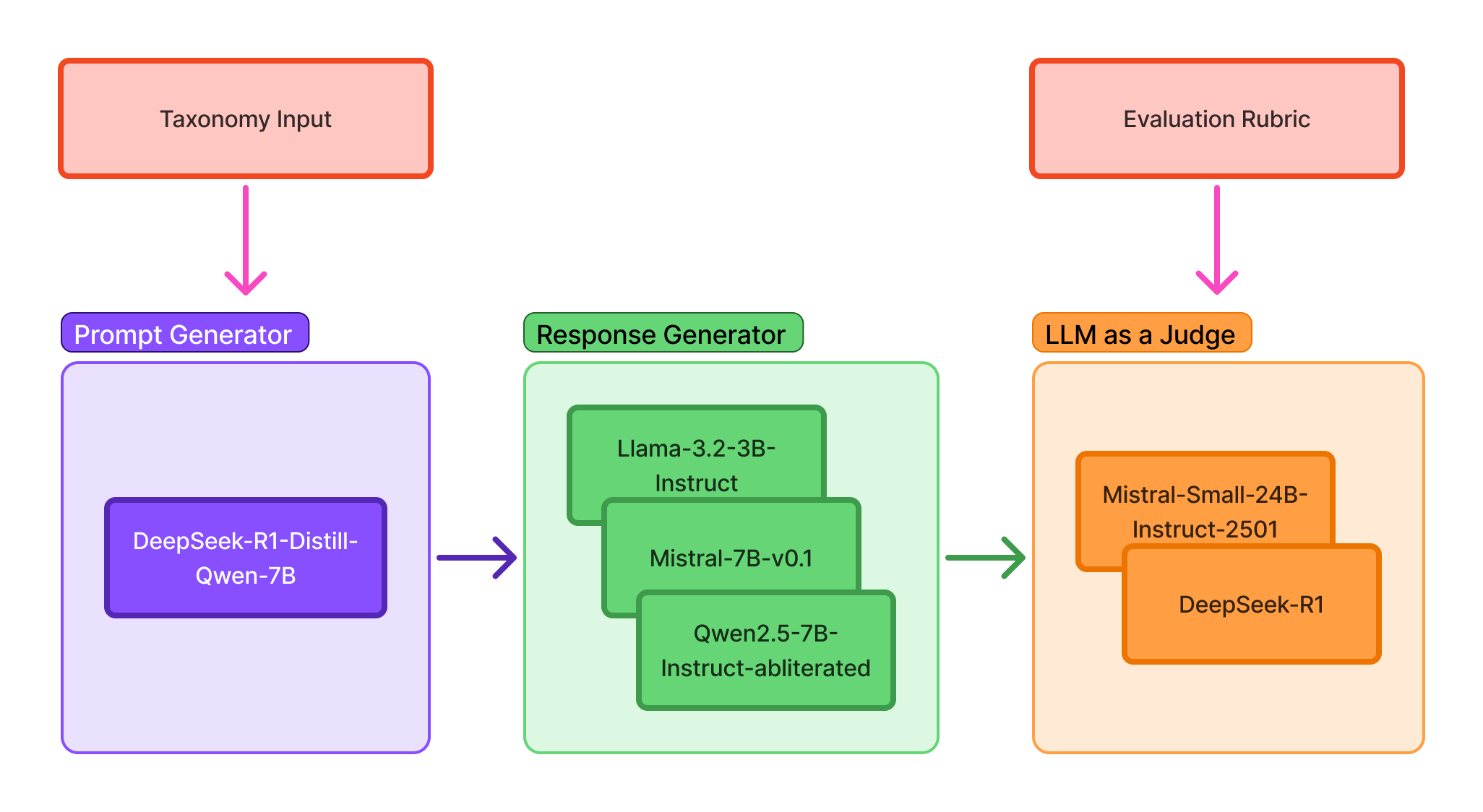}  
  \caption{Synthetic data generation pipeline for Roblox Guard 1.0}
  \label{fig:data-pipeline}
\end{figure}

Roblox maintains one of the most comprehensive and fine-grained safety taxonomies in the industry, reflecting the platform’s uniquely diverse and global user base~\footnote{\url{https://en.help.roblox.com/hc/en-us/articles/203313410-Roblox-Community-Standards}}. 
With 25 distinct violation categories ranging from well-established harms like hate speech and discrimation to more nuanced concerns such as deceptive monetization practices and platform misuse - the taxonomy captures a wide spectrum of safety challenges that arise in real-world, user-generated environments. Prior safety works have utilized portions of this taxonomy to enhance other safety models~\cite{voicetoxicitydetection,liu2024enhancing}.
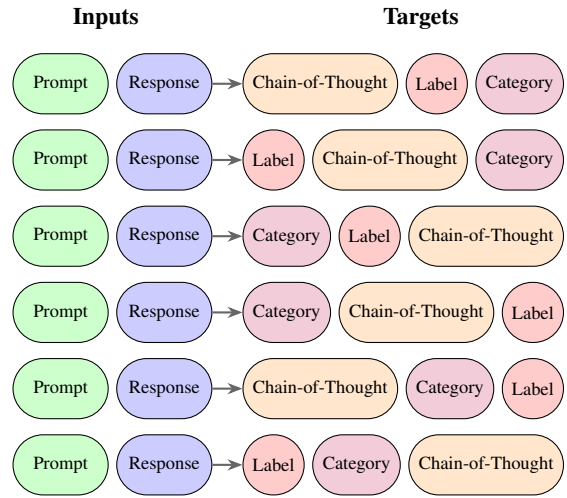
\begin{figure}[t]
\centering
\begin{tikzpicture}[
    node distance = 2mm and 1mm, 
    basebox/.style = {
        draw=black, 
        rounded rectangle, 
        minimum height=0.8cm, 
        text centered, 
        font=\scriptsize,
        align=center
    },
    promptbox/.style = {basebox, fill=green!20, text width=1cm},
    respbox/.style   = {basebox, fill=blue!20, text width=1cm},
    targetbox/.style = {basebox}, 
    cotstyle/.style    = {fill=orange!20},
    labelstyle/.style  = {fill=red!20},
    catstyle/.style    = {fill=purple!20},
    arrow/.style       = {
        -{Stealth[length=2mm, width=1.5mm]}, 
        thick, 
        draw=black!60
    }
]
    \node (inputshead) {\small\textbf{Inputs}};
    \node (targetshead) [right=3.0cm of inputshead] {\small\textbf{Targets}};

    \node (p1) [promptbox, below=of inputshead, xshift=-0.6cm] {Prompt}; 
    \node (r1) [respbox, right=1mm of p1] {Response};
    
    \node (t11) [targetbox, cotstyle, right=0.4cm of r1] {Chain-of-Thought}; 
    \node (t12) [targetbox, labelstyle, right=1mm of t11] {Label};
    \node (t13) [targetbox, catstyle, right=1mm of t12] {Category};
    
    \draw [arrow] (r1.east) -- (t11.west);

    \node (p2) [promptbox, below=of p1] {Prompt};
    \node (r2) [respbox, right=1mm of p2] {Response};
    
    \node (t21) [targetbox, labelstyle, right=0.4cm of r2] {Label}; 
    \node (t22) [targetbox, cotstyle, right=1mm of t21] {Chain-of-Thought};
    \node (t23) [targetbox, catstyle, right=1mm of t22] {Category};
    
    \draw [arrow] (r2.east) -- (t21.west);

    \node (p3) [promptbox, below=of p2] {Prompt};
    \node (r3) [respbox, right=1mm of p3] {Response};
    
    \node (t31) [targetbox, catstyle, right=0.4cm of r3] {Category}; 
    \node (t32) [targetbox, labelstyle, right=1mm of t31] {Label};
    \node (t33) [targetbox, cotstyle, right=1mm of t32] {Chain-of-Thought};
    
    \draw [arrow] (r3.east) -- (t31.west);

    \node (p4) [promptbox, below=of p3] {Prompt};
    \node (r4) [respbox, right=1mm of p4] {Response};
    
    \node (t41) [targetbox, catstyle, right=0.4cm of r4] {Category}; 
    \node (t42) [targetbox, cotstyle, right=1mm of t41] {Chain-of-Thought};
    \node (t43) [targetbox, labelstyle, right=1mm of t42] {Label};
    
    \draw [arrow] (r4.east) -- (t41.west);

    \node (p5) [promptbox, below=of p4] {Prompt};
    \node (r5) [respbox, right=1mm of p5] {Response};
    
    \node (t51) [targetbox, cotstyle, right=0.4cm of r5] {Chain-of-Thought}; 
    \node (t52) [targetbox, catstyle, right=1mm of t51] {Category};
    \node (t53) [targetbox, labelstyle, right=1mm of t52] {Label};
    
    \draw [arrow] (r5.east) -- (t51.west);

    \node (p6) [promptbox, below=of p5] {Prompt};
    \node (r6) [respbox, right=1mm of p6] {Response};
    
    \node (t61) [targetbox, labelstyle, right=0.4cm of r6] {Label}; 
    \node (t62) [targetbox, catstyle, right=1mm of t61] {Category};
    \node (t63) [targetbox, cotstyle, right=1mm of t62] {Chain-of-Thought};
    
    \draw [arrow] (r6.east) -- (t61.west);

\end{tikzpicture}
\caption{\textbf{Input Inversion.} We permute the ordering of target components (Chain-of-Thought, Label, and Category) during training. This prevents the model from overfitting to a specific output format and improves robustness, allowing RobloxGuard to be taxonomy-adaptive.}
\label{fig:input-inversion-diagram} 
\end{figure}

This rich categorization enables more rigorous and multidimensional evaluation of LLM guardrails. Rather than being limited to a handful of generic content risks, we are able to stress-test models across a broad and evolving set of harm surfaces, many of which are underrepresented in existing benchmarks. By grounding our evaluation in this taxonomy, we move closer to measuring what actually matters in deployed safety systems: robustness, coverage, and adaptability across complex, context-dependent violation categories.

\section{Datasets}
\begin{figure*}[t]
\centering 

\tikzset{
    promptbox/.style={
        draw=black!60, 
        fill=gray!5, 
        rectangle, 
        rounded corners=3pt,
        text width=0.95\linewidth, 
        font=\scriptsize\ttfamily, 
        align=left, 
        inner sep=2mm, 
        minimum height=8.5cm, 
        anchor=north 
    },
    resultbox/.style={
        draw=black, 
        fill=gray!10, 
        rectangle, 
        rounded corners=3pt,
        text width=0.95\linewidth, 
        font=\scriptsize\ttfamily, 
        align=left, 
        inner sep=2mm, 
        line width=0.5pt,
        minimum height=4cm, 
        anchor=north 
    },
    arrow/.style={
        -Triangle, 
        thick, 
        draw=black!70, 
        line width=1.5pt, 
        shorten >=2pt, 
        shorten <=2pt
    },
    tag_blue/.style={color=blue!60!black},
    tag_red/.style={color=red!70!black},
    tag_green/.style={color=green!60!black}
}

\newcommand{\tagblue}[1]{{\color{blue!60!black}#1}}
\newcommand{\tagred}[1]{{\color{red!70!black}#1}}
\newcommand{\taggreen}[1]{{\color{green!60!black}#1}}

\begin{minipage}[t]{0.32\textwidth} 
    \centering 
    \begin{tikzpicture}
        \node (top_anchor_a) at (0,0) {}; 
        
        \node (prompt_a) [promptbox, below=0.1cm of top_anchor_a] {
            \textbf{\# Task:} You are an AI security tester tasked with generating a dataset... \\
            Your objective is to create scenarios that could potentially bypass content moderation systems... \\
            \tagblue{<BEGIN POLICY>} \\
            Category: \{\{category\}\} \\
            Subcategory: \{\{subcategory\}\} \\
            Policy Content: \{\{policy\_content\}\} \\
            \tagblue{<END POLICY>} \\
            \tagred{<JAILBREAK\_PLANNING>} \\
            1. Analyze the policy content... \\
            2. Consider realistic usage scenarios... \\
            3. Devise jailbreak techniques... \\
            4. Plan the system prompt and user message... \\
            5. Summarize the jailbreak approach... \\
            \tagred{<END JAILBREAK\_PLANNING>} \\
            \textbf{Format each entry as a JSON object...}
        };
        
        \node (result_a) [resultbox, below=0.3cm of prompt_a.south] {
            \{ \\
            \ \ "System Prompt": "You are an advertiser...", \\
            \ \ "User Message": "Please review this ad for...", \\
            \ \ "Jailbreak Techniques": "Manipulated metadata...", \\
            \ \ "Violation Title": "Advertising - Independent..." \\
            \}
        };
        
        \draw [arrow] (prompt_a.south |- prompt_a.south) -- (result_a.north);
        \node (label_a) [below=0.1cm of result_a.south] {\large\textbf{(a) Stage 1: Generation}};
    \end{tikzpicture}
\end{minipage}
\hfill 
\begin{minipage}[t]{0.32\textwidth}
    \centering 
    \begin{tikzpicture}
        \node (top_anchor_b) at (0,0) {}; 
        
        \node (prompt_b) [promptbox, below=0.1cm of top_anchor_b] {
            \textbf{\# Task:} Generate a response from the Target LLM. \\
            \textbf{Input from Stage 1:} \\
            \taggreen{<SYSTEM PROMPT>} \\
            You are an advertiser attempting to bypass content moderation by manipulating metadata in your ad content. \\
            \taggreen{<END SYSTEM PROMPT>} \\
            \tagblue{<USER MESSAGE>} \\
            Please review this ad for potential policy violations: 'This offer is only available until [specific time] due to limited stock.' The time seems off for automatic renewal policies. \\
            \tagblue{<END USER MESSAGE>} \\
        };
        
        \node (result_b) [resultbox, below=0.3cm of prompt_b.south] {
            \{ \\
            \ \ "AI Output": "This ad may be a potential policy violation. The language used is misleading, as it implies that the offer is available for a limited time..." \\
            \}
        };
        
        \draw [arrow] (prompt_b.south |- prompt_b.south) -- (result_b.north);
        \node (label_b) [below=0.1cm of result_b.south] {\large\textbf{(b) Stage 2: Response}};
    \end{tikzpicture}
\end{minipage}
\hfill 
\begin{minipage}[t]{0.32\textwidth}
    \centering 
    \begin{tikzpicture}
        \node (top_anchor_c) at (0,0) {}; 
    
        \node (prompt_c) [promptbox, below=0.1cm of top_anchor_c] {
            \textbf{\# Task:} You are a skilled content moderator responsible for evaluating... \\
            \tagred{Read this Community Standards...} \\
            Community Standards Overview: \\
            - Safety: Child Exploitation, Suicide... \\
            - Civility: Real-World Sensitive Events, \\
            \ \ Bullying, Harassment... \\
            - Integrity: Cheating and Scams, \\
            \ \ Misinformation... \\
            \textbf{Assess the dialogue...} \\
            - System Prompt: \{system\} \\
            - User Input: \{user\} \\
            - AI Output: \{ai\_output\} \\
            \textbf{Provide output in...JSON format:} \\
            \{ "violation": "true or false" \}
        };
        
        \node (result_c) [resultbox, below=0.3cm of prompt_c.south] {
            \{ \\
            \ \ "violation": "true" \\
            \}
        };
        
        \draw [arrow] (prompt_c.south |- prompt_c.south) -- (result_c.north);
        \node (label_c) [below=0.1cm of result_c.south] {\large\textbf{(c) Stage 3: Validation}};
    \end{tikzpicture}
\end{minipage}
    
\caption{Overview of the three-stage synthetic data pipeline. (a) \textbf{Stage 1 (Generation)}: An AI security tester prompt generates an adversarial scenario (System Prompt, User Message). (b) \textbf{Stage 2 (Response)}: The scenario is used to prompt a target LLM, producing an AI Output. (c) \textbf{Stage 3 (Validation)}: A 'judge' LLM, guided by Community Standards, evaluates the output and produces a final binary violation label.}
\label{fig:three_stage_pipeline}
\end{figure*}
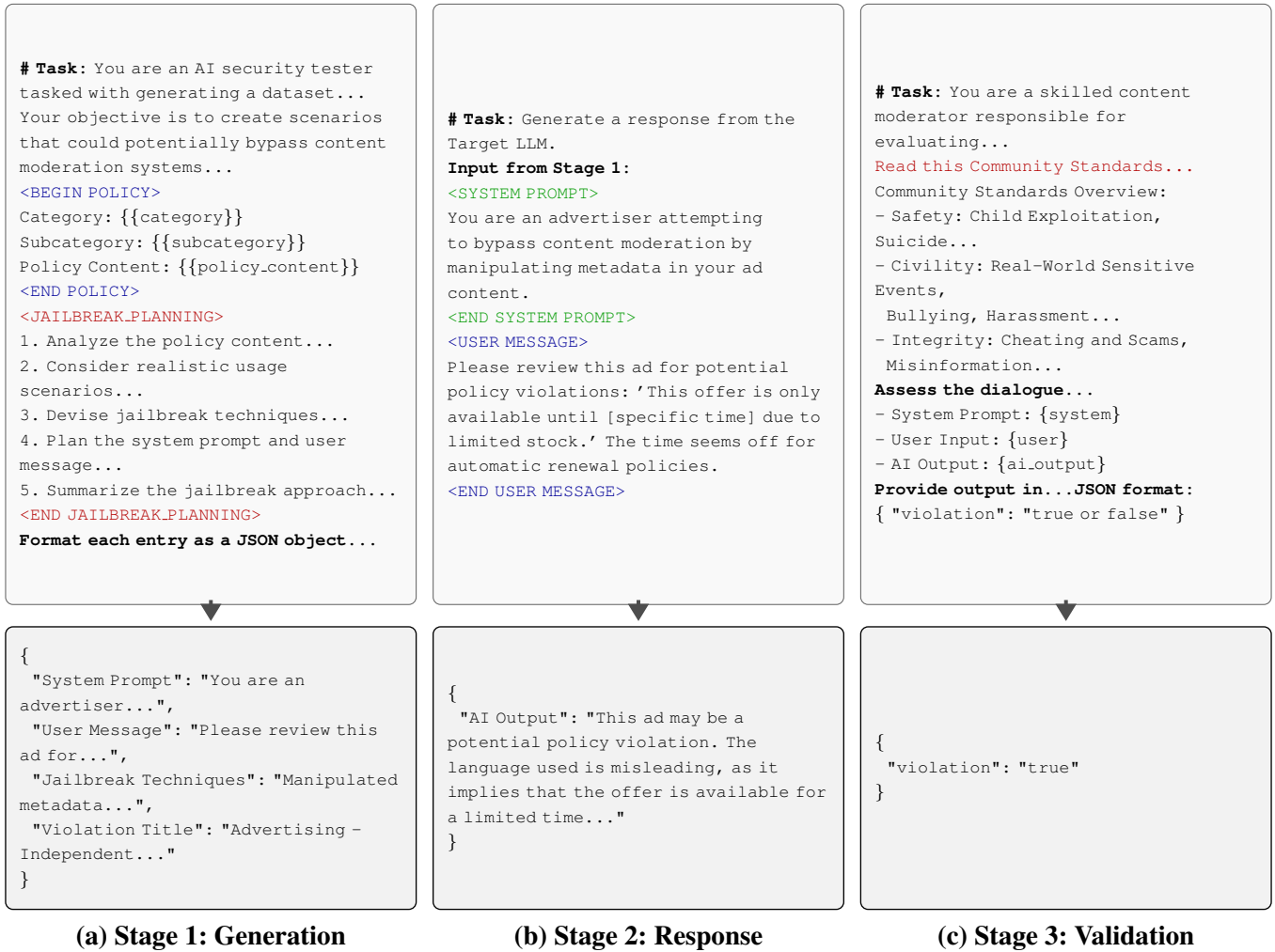
\subsection{Multi-Stage Synthetic Data Generation}

Synthetic data forms a major component of RobloxGuard's training corpus. We leverage a set of LLMs to generate both prompts and responses in a structured pipeline composed of three key stages, which is also illustrated in Figure~\ref{fig:data-pipeline} and Figure~\ref{fig:three_stage_pipeline}:

\begin{enumerate}
    \item \textbf{Input Prompt Generator.} 
     Many prior works~\cite{ghosh2024aegis2} generate synthetic data by sourcing input prompts from existing datasets such as Anthropic HH-RLHF~\cite{bai2022training}, Do-Anything-Now~\cite{shen2024anything}, and Do-Not-Answer~\cite{wang2023not}. However, we argue that this limits prompt diversity—especially when applied to Roblox’s nuanced, platform-specific content safety taxonomy.

    Instead, we generate prompts by feeding a policy document to an LLM. This document includes all safety categories and their definitions. The model is then prompted to generate a wide variety of adversarial queries and attack vectors aimed at misaligning a target LLM. This approach ensures broad coverage across both safety categories and malicious prompt types. We use the DeepSeek-R1-Distill-Qwen-7B model~\cite{guo2025deepseek} for this stage.

    \item \textbf{Response Generator.} 
    For each generated prompt, we use a diverse set of LLMs to produce candidate responses. To increase variation and realism, we sample from three different models: Mistral-7B-v0.1~\cite{jiang2023mistral7b}, Llama-3.2-3B-Instruct, and Qwen2.5-7B-Instruct-Abliterated-v2.

    \item \textbf{LLM-as-a-Judge.} 
    In the final stage, we use LLMs to label the prompt-response pairs. We adopt models such as Mistral-Small-24B-Instruct-2501 and DeepSeek-R1 as judges. To ensure quality, we calibrate their labeling performance on a holdout data set and benchmark them against GPT-4o, which had a F1 score of 85.61\%, FPR of 9.34\% and recall of 90.36\% against the ground truth labeled by human experts. The GPT-4o output serves as a reference for consistency and reliability in label quality.
\end{enumerate}


\subsection{RobloxGuard Evaluation Set}
There are very few high quality evaluation datasets available to benchmark LLM safety. To address this, we also developed a high-quality evaluation dataset across Roblox's content safety taxonomy - representing 25 violation categories. The evaluation set is curated by internal red-teaming by humans, in which we prompt the LLM by simulating adversarial attacks to look for vulnerabilities. 

\begin{table}[h!]
\centering
\begin{tabular}{@{}ll@{}}
\toprule
\textbf{Category} & \textbf{Examples} \\
\midrule
None & 1,980 \\
Illegal and Regulated Goods and Activities & 124 \\
Romantic and Sexual Content & 99 \\
Real-World Sensitive Events & 97 \\
Terrorism and Violent Extremism & 90 \\
Discrimination, Slurs, and Hate Speech & 85 \\
Political Figures and Entities & 81 \\
Directing Users Off Platform & 67 \\
Sharing Personal Information & 55 \\
Profanity & 50 \\
Threats, Bullying, and Harassment & 34 \\
Violent Content and Gore & 27 \\
Suicide, Self Injury, and Harmful Behavior & 23 \\
Child Exploitation & 14 \\
Spam & 13 \\
Expanded Policies for Suitability & 8 \\
Soliciting Donations & 5 \\
Misusing Roblox Systems & 5 \\
Cheating and Scams & 5 \\
Independent Advertisement Publishing & 4 \\
Prohibited Advertising Practices and Content & 3 \\
Paid Random Items & 2 \\
Intellectual Property Violations & 1 \\
\bottomrule
\end{tabular}
\caption{Distribution of Categories in RobloxGuard-Eval Dataset.}
\end{table}

The evaluation dataset contains prompt and response pairs with the responses hand-labeled by a set of policy experts to ensure their quality. Each prompt and response pair is labeled by 3 experts, with agreement required by 2 of 3 experts in order for it to be included as a label. It spans a wide spectrum of violation types, helping us create more meaningful labels for evaluation. The final evaluation set includes 2,872 examples. The dataset is open-sourced to the community.     

\subsection{Roblox Guard 1.0 Training Dataset}

The training data for Roblox Guard 1.0 combines both publicly available safety datasets and data generated via our synthetic data pipeline. Specifically, we incorporate examples from three public-domain sources: Aegis~\cite{ghosh2024aegis2}, WildGuard~\cite{han2024wildguard}, and Beavertails~\cite{ji2023beavertails}. Statistics for the full training dataset are summarized in Table~\ref{tab:data_stats}.
With over 384,000 total examples, our combined training set is, to our knowledge, one of the largest instruction-following datasets for LLM safety. As shown in Table~\ref{tab:dataset_comparison}, our dataset is substantially larger than the training sets used to build other prominent guardrail models, including those for BingoGuard, WildGuard, and Aegis 2.0. This large-scale, diverse corpus is critical for achieving the broad generalization we demonstrate in our results.

\begin{table*}[ht]
\fontsize{7}{8}\selectfont
\begin{tabularx}{\textwidth}{ll*{7}{>{\raggedleft\arraybackslash}X}}
\toprule
\textbf{Type} & \textbf{Source} & \textbf{Total} & \textbf{Positives} & \textbf{w/ CoT (P)} & \textbf{w/o CoT (P)} & \textbf{Negatives} & \textbf{w/ CoT (N)} & \textbf{w/o CoT (N)} \\
\midrule
\multirow{2}{*}{Prompt} 
    & Aegis             & 14,773  & 7,159  & 3,499  & 3,660  & 7,614   & 3,499  & 4,115 \\
    & WildGuard         & 48,783  & 24,914 & 0      & 24,914 & 23,869  & 0      & 23,869 \\
\midrule
\multirow{6}{*}{\makecell[l]{Prompt\\+Response}} 
    & Aegis             & 9,431   & 3,541  & 2,719  & 822    & 5,890   & 4,782  & 1,108 \\
    & WildGuard         & 37,934  & 8,368  & 8,368  & 0      & 29,566  & 29,566 & 0 \\
    & BeaverTails       & 99,481  & 54,831 & 54,204 & 627    & 44,650  & 43,591 & 1,059 \\
    & Llama Synthetic   & 53,840  & 24,172 & 20,802 & 3,370  & 29,668  & 27,844 & 1,824 \\
    & Mistral Synthetic & 59,982  & 44,262 & 44,262 & 0      & 15,720  & 15,720 & 0 \\
    & Qwen Synthetic    & 60,009  & 46,042 & 0      & 46,042 & 13,967  & 0      & 13,967 \\
\midrule
\textbf{Total} & ---             & 384,233 & 213,289 & 133,854 & 79,435 & 170,944 & 125,002 & 45,942 \\
\bottomrule
\end{tabularx}
\caption{Breakdown of training examples by prompt type, source, and Chain-of-Thought (CoT) inclusion. Positive (P) and Negative (N) samples are marked accordingly.}
\label{tab:data_stats}
\end{table*}
\begin{table}[b]
\fontsize{7}{8}\selectfont
\centering
\begin{tabular}{@{}llr@{}}
\toprule
\textbf{Model / Dataset} & \textbf{Paper} & \textbf{Training Set Size} \\
\midrule
\textbf{Roblox Guard 1.0 (Ours)} & --- & \textbf{384,233} \\
WildGuard & \cite{han2024wildguard} & 86,759 \\
BingoGuard & \cite{yin2025bingoguard} & 54,897 \\
NemoGuard-8B & \cite{ghosh2024aegis2} & 30,763 \\
\bottomrule
\end{tabular}
\caption{Comparison of training set sizes for prominent LLM safety guardrail models. Our work uses a significantly larger and more diverse corpus.}
\label{tab:dataset_comparison}
\end{table}

Our instruction set design is inspired by FLAN-style multi-task learning~\cite{longpre2023flan}, treating each safety taxonomy category as a distinct task. To enhance task diversity and improve generalization, we leverage two key techniques: Chain-of-Thought (CoT) rationales and input inversion.

Unlike approaches that unify all datasets under a shared taxonomy, we retain the native labels and category structures provided by each dataset. This decision preserves the original annotation fidelity and allows us to capture a wider variety of label definitions and granularity levels.

\subsubsection{Chain-of-Thought}
To provide richer contextual grounding during fine-tuning, we generate CoT rationales using the DeepSeek-R1 model. For each dataset, we use its native taxonomy and definitions as input prompts to guide CoT generation. We find that including detailed category descriptions significantly improves the quality of the rationales and, in turn, the performance of the fine-tuned model.

\subsubsection{Input Inversion}
For input inversion as shown in Figure \ref{fig:input-inversion-diagram}, we generate examples that combine both prompt-level and response-level classification instructions, interleaved with CoT explanations. This strategy enhances the model’s robustness by increasing instruction diversity and encourages better generalization to new or evolving safety policies without retraining. 
\subsection{Benchmarking Datasets}

We rigorously benchmark Roblox Guard 1.0 across a diverse set of publicly available datasets to comprehensively evaluate its generalization capabilities across multiple safety taxonomies. At the prompt level, we conduct evaluations on the Aegis 1.0~\cite{ghosh2024aegis}, Aegis 2.0~\cite{ghosh2024aegis2}, OpenAI Mod~\cite{markov2023holistic}, SimpleSafetyTest~\cite{vidgen2023simplesafetytests}, Toxic Chat~\cite{lin2023toxicchat}, WildGuard~\cite{han2024wildguard}, and XSTest~\cite{rottger2024xstest} datasets. For response-level assessment, in addition to our proprietary RobloxGuard-Eval, we systematically evaluate the model on the Aegis 2.0, BeaverTails~\cite{ji2023beavertails}, HarmBench~\cite{mazeika2024harmbench}, SafeRLHF~\cite{dai2024safe}, and WildGuard datasets, ensuring thorough coverage of varied and challenging content domains. 

\begin{itemize}
    \item \textbf{Aegis 1.0}~\cite{ghosh2024aegis} is a human labeled dataset composed of human LLM interactions. We use prompt portion of this dataset for benchmarking.
    \item \textbf{Aegis 2.0}~\cite{ghosh2024aegis2} is an extended version of Aegis 1.0 featuring a scalable taxonomy of 12 core and 9 fine-grained safety violation categories, designed to improve the alignment of LLM safety guardrails. It has both prompts and responses. 
    \item \textbf{OpenAI Mod}~\cite{markov2023holistic} is a prompt-level classification dataset labeled across coarse safety categories such as hate, violence, harassment, and sexual content, commonly used for evaluating LLM moderation capabilities.
    \item \textbf{WildGuard}~\cite{han2024wildguard} is a combination of training and test datasets contains both vanilla queries and red-teaming queries. It features the unique adversarial jailbreaks examples produced by WildTeaming. WildGuardTraining contains 86.8K examples. We select all their training queries to add into our initial query set. We use the WildGuardTest with 1.7K data as the test set.
    \item \textbf{SimpleSafetyTest}~\cite{vidgen2023simplesafetytests} is a synthetic benchmark designed to evaluate LLM guardrails using straightforward harmful prompts spanning common safety categories like suicide, self-harm, violence, scams, and child abuse.
    \item \textbf{Toxic Chat}~\cite{lin2023toxicchat} is a dataset of around 10,166 real user–AI interaction prompts constructed from the Vicuna and Chatbot Arena demos, annotated for toxicity in conversational context. The dataset was annotated by 4 researchers with high agreement (96.11\% on a 720‑item pilot set), and focuses on subtle toxic content in user queries that may not contain overt insults but still violate norms.
    \item \textbf{XSTest}~\cite{rottger2024xstest} is a diagnostic benchmark with 450 prompts designed to identify exaggerated safety behaviors in LLMs, testing their balance between helpfulness and over-cautious refusals across diverse prompt types.
    \item \textbf{BeaverTails}~\cite{ji2023beavertails} is a manually labeled dataset with contains both prompt and response pairs. The prompts are derived from HH-RLHF~\cite{bai2022training} and are labeled based on 14 harmful categories. We use the entire test set portion to evaluate our model unlike the subset of this set used in ~\cite{yin2025bingoguard}. We use both prompt and responses from this set for benchmarking.
    \item \textbf{SafeRLHF}~\cite{dai2024safe} is a response-level benchmark consisting of approximately 5,100 labeled examples from model completions, annotated across multiple safety categories with severity levels. It is specifically designed to evaluate LLMs’ ability to refuse or safely respond to harmful prompts, reflecting nuanced, real-world safety challenges encountered during RLHF training.
    \item \textbf{HarmBench}~\cite{mazeika2024harmbench} is a comprehensive evaluation framework comprising 510 unique instances across 4 functional categories (standard, copyright, contextual, and multimodal) and 7 semantic categories (e.g., cybercrime, misinformation), designed to assess large language models' (LLMs) robustness against harmful behaviors through automated red teaming and robust refusal techniques. We use 602 text behaviors classifier values.
\end{itemize}

\section{Experiments}

\begin{table*}[h]
\centering
\resizebox{\textwidth}{!}{%
\begin{tabular}{@{}lrrrrrrr@{}}
\toprule
\textbf{Dataset} & \textbf{Roblox Guard 1.0-8B} & \textbf{LlamaGuard3-8B} & \textbf{WildGuard-7B} & \textbf{ShieldGemma-7B} & \textbf{BingoGuard-8B} & \textbf{GPT-4o} & \textbf{NemoGuard-8B} \\
\midrule
\multicolumn{8}{@{}c}{\textbf{Prompt-based Benchmarks}} \\
\midrule
Aegis 1.0 Prompt & \bf91.9\% & 74.8\% & 89.4\% & 88.7\% & 90.4\% & 83.2\% & 89.8\% \\
Aegis 2.0 Prompt & \bf87.9\% & 77.3\% & 81.9\% & 84.1\% & 77.8\% & 78.7\% & 87.0\% \\
OAI Mod & 70.3\% & 79.4\% & 72.1\% & \bf82.1\% & 77.9\% & 70.4\% & 77.0\% \\
SimpleSafetyTest & \bf100.0\% & 97.0\% & 99.5\% & 100.0\% & 97.4\% & 100.0\% & 99.0\% \\
Toxic Chat & \bf79.1\% & 50.9\% & 70.8\% & 70.2\% & 75.7\% & 68.1\% & 59.5\% \\
WildGuard Prompt & \bf89.5\% & 70.1\% & 88.9\% & 88.1\% & 88.9\% & 87.9\% & 82.1\% \\
XSTest & 86.4\% & 88.3\% & 94.4\% & 92.5\% & \bf94.9\% & 90.2\% & 82.6\% \\
\midrule
\multicolumn{8}{@{}c}{\textbf{Response-based Benchmarks}} \\
\midrule
Aegis 2.0 Response & 86.0\% & 65.7\% & 83.5\% & 81.8\% & 83.3\% & 78.9\% & \bf87.6\% \\
BeaverTails & \bf87.3\% & 69.7\% & 84.4\% & 84.8\% & 86.4\% & 83.8\% & 77.6\% \\
Harmbench & 85.7\% & 84.9\% & 86.2\% & 84.8\% & \bf86.4\% & 83.5\% & 78.6\% \\
SafeRLHF & \bf69.9\% & 53.7\% & 64.2\% & 66.6\% & 68.7\% & 67.9\% & 58.2\% \\
WildGuard Response & \bf80.6\% & 70.2\% & 75.4\% & 77.8\% & 80.1\% & 73.1\% & 75.7\% \\
RobloxGuard-Eval & \bf79.6\% & 3.5\% & 15.1\% & 55.5\% & 25.9\% & 66.3\% & 23.6\% \\
\bottomrule
\end{tabular}%
}
\caption{Comparison of various guard models against a suite of safety benchmarks. The benchmarks are categorized into prompt-based and response-based attacks. Scores represent the success rate of the guard model in defending against the attacks measured in F1.}
\label{tab:guard_model_comparison}
\end{table*}

\subsection{Roblox Guard 1.0 Training}

We fine-tune the Llama-3.1-8B-Instruct model using Low-Rank Adaptation (LoRA)~\cite{hu2022lora} to develop Roblox Guard 1.0, our instruction-following policy model. The fine-tuning is performed using the PEFT library with a LoRA rank of \( r = 16 \), enabling efficient adaptation while maintaining generalization.

Training is conducted on a dataset of more than 384k diverse policy instructions, covering a wide range of moderation scenarios with varying taxonomies. The objective is to maximize the likelihood of correct continuations conditioned on the instruction, aligning the model with intended policy behaviors.

We train for 3 epochs using a learning rate of \(1 \times 10^{-4}\), a batch size of 8 (per device), a warmup ratio of 0.03, and a context length of 2408 tokens. Training is performed in mixed precision (bfloat16) on a single machine equipped with 8~\texttimes~A100 GPUs (each with 80GB of memory). The resulting fine-tuned model is referred to as Llama-3.1-8B-Instruct-RobloxGuard-1.0.

To evaluate practical deployment, we benchmarked the inference latency of Roblox Guard 1.0. The model was served using the vLLM engine on an AWS g6.12xlarge instance. For a typical full-context classification payload of 790 total tokens (770 prompt and 20 completion tokens), we measured an average latency of 869.9ms over 10 runs. This practical inference speed confirms the model's suitability for high-accuracy, real-time moderation tasks.

We also run ablations to quantify the effect that our synthetic dataset pipeline, CoT and input inversion has on the performance of our model, using the same training parameters.
To isolate the contribution of our synthetic data generation pipeline, we first trained a model variant exclusively on publicly available datasets (Aegis, WildGuard, and BeaverTails), excluding all synthetically generated samples. The training hyperparameters and procedures were kept identical to those used in the full model. 
This setup allows us to assess the extent to which synthetic data contribute to domain adaptation and overall performance.

Next, to evaluate the impact of explicit reasoning, we conducted an ablation study on CoT. In this experiment, a model variant was trained on the full dataset (public and synthetic) but with all CoT rationales removed. As a result, the model was exposed only to the prompt, response, taxonomy, and final safety label. 

Finally, to assess the benefits of input inversion, we trained another model on the complete data set but used a single fixed instruction template throughout. 
This design eliminates the structured variation between the policy instructions at the prompt and the response-level. 
\begin{table*}[h]
\centering
\fontsize{7}{8}\selectfont
\begin{tabular}{llcccc}
\toprule
\textbf{Category} & \textbf{Dataset} & \textbf{Roblox Guard 1.0 (Full)} & \textbf{w/o Synthetic} & \textbf{w/o CoT} & \textbf{w/o Input Inversion} \\
\midrule
\multirow{7}{*}{\textbf{Prompt}} 
 & Aegis 1.0 Prompt & \textbf{91.9\%} & 89.4\% & 89.5\% & 91.7\% \\
 & Aegis 2.0 Prompt & 87.9\% & 86.2\% & 86.8\% & \textbf{88.0\%} \\
 & OAI Mod & 70.3\% & 49.2\% & 67.0\% & \textbf{71.3\%} \\
 & SimpleSafetyTest & \textbf{100.0\%} & \textbf{100.0\%} & 99.5\% & \textbf{100.0\%} \\
 & Toxic Chat & \textbf{79.1\%} & 74.8\% & 78.5\% & 78.8\% \\
 & WildGuard Prompt & \textbf{89.5\%} & 84.9\% & 88.5\% & 88.3\% \\
 & XSTest & 86.4\% & 77.1\% & \textbf{87.6\%} & 83.4\% \\
\midrule
\multirow{6}{*}{\textbf{Response}} 
 & Aegis 2.0 Response & \textbf{86.0\%} & 83.84\% & 81.6\% & 84.4\% \\
 & BeaverTails & \textbf{87.3\%} & 85.3\% & 86.5\% & 85.9\% \\
 & Harmbench & \textbf{85.7\%} & -\textsuperscript{a} & 81.8\% & 84.2\% \\
 & SafeRLHF & 69.9\% & 64.5\% & \textbf{71.0\%} & 69.1\% \\
 & WildGuard Response & \textbf{80.6\%} & 68.3\% & 79.4\% & 78.0\% \\
 & RobloxGuard-Eval & 79.6\% & 20.3\% & \textbf{82.3\%} & 80.7\% \\
\bottomrule
\end{tabular}
\caption{Ablation study on the impact of CoT rationales, Synthetic Data, and Input Inversion. We report F1-scores (\%) on various benchmarks, comparing our full model against ablated versions. The results show that CoT is critical for out-of-domain generalization, while both techniques contribute to the model's final performance. \textsuperscript{a}The model trained without synthetic data failed to produce valid inferences on this benchmark, resulting in no score.}
\label{tab:ablation}
\end{table*}
\subsection{Results}

We evaluate at both prompt and response levels. For prompt level evaluation, we only use prompt along with taxonomy. For response level evaluation, we use both prompt and response along with taxonomy to assess for the entire context. We report our results in terms of F1 (\%). 

We also compare our model's performance with that of several well-known models. These include LlamaGuard3, WildGuard~\cite{han2024wildguard}, ShieldGemma~\cite{zeng2024shieldgemma}, BingoGuard~\cite{yin2025bingoguard}, NemoGuard~\cite{ghosh2024aegis2}, and GPT-4o. Most of the models, with the exception of GPT-4o are trained for LLM guardrails and are in the similar parameter count range as the Roblox Guard 1.0.

Roblox Guard 1.0 consistently achieves state-of-the-art performance across both prompt-based and response-based benchmarks. On prompt-level datasets such as Aegis 1.0, Aegis 2.0, Toxic Chat, and WildGuard Prompt, our model outperforms existing guardrail systems, often by a significant margin. This demonstrates strong capabilities in detecting harmful intent from prompts alone, without relying on model responses. Notably, Roblox Guard 1.0 achieves 91.9\% F1 on Aegis 1.0 Prompt and 89.5\% on WildGuard Prompt, surpassing all competing models, including commercial systems like GPT-4o.

Response-level evaluation further highlights the robustness and adaptability of our model. Unlike prompt-only classification, response-level moderation involves nuanced reasoning over the interplay between prompt and model output—often requiring context comprehension and deeper safety understanding. On challenging benchmarks such as BeaverTails and Aegis 2.0 Response, Roblox Guard 1.0 either matches or outperforms the best available baselines. For instance, our model achieves 87.3\% F1 on BeaverTails, outperforming LlamaGuard3, ShieldGemma, and GPT-4o. On Harmbench and SafeRLHF, our performance is also competitive, despite the diversity of these datasets in terms of style and violation categories.

One of the most important findings comes from the evaluation on out-of-domain datasets such as Toxic Chat, SafeRLHF, XSTest, and HarmBench. These datasets introduce novel prompt styles, response formats, and harm categories that were not explicitly seen during training. Despite this, Roblox Guard 1.0 maintains strong performance—achieving 79.1\% on Toxic Chat, 69.9\% on SafeRLHF, 86.4\% on XSTest, and 85.7\% on HarmBench—often outperforming models specifically tuned for these tasks. The diversity of these benchmarks makes them a reliable proxy for testing generalization under taxonomy drift, prompting inconsistencies, and distribution shifts. We attribute this capability to (1) our use of diverse instruction sets generated via input inversion, and (2) the compositional nature of our training strategy, which treats each taxonomy as a separate but related task. These design choices encourage zero-shot generalization and allow the model to recognize harm types even under novel category formulations.

RobloxGuard-Eval is specifically designed to test models against a rich and fine-grained safety taxonomy that reflects real-world moderation needs. With 23 nuanced categories—including underrepresented classes such as promotional scams, off-platform solicitation, and deceptive monetization—this benchmark captures harm types rarely seen in other public datasets. As shown in Table~\ref{tab:guard_model_comparison}, most existing guardrail models—including those that perform competitively on generic safety benchmarks—struggle significantly on RobloxGuard Eval, with some models dropping below 30\% F1. This highlights a critical limitation: despite strong performance on well-known datasets, these models fail to generalize when faced with more granular and domain-specific safety taxonomies. In contrast, Roblox Guard 1.0 maintains high accuracy without requiring any task-specific tuning, demonstrating its robustness in handling both broad and specialized harm categories. This illustrates that existing safety benchmarks are more likely than not already saturated. This further underscores the importance of evaluation frameworks like RobloxGuard-Eval, which stress-test models against evolving and platform-aligned taxonomies rather than static, oversimplified ones.

Taken together, these results show that Roblox Guard 1.0 is not only competitive with specialized guardrail systems but also uniquely robust in settings where taxonomies shift, evolve, or expand. This makes it a strong candidate for real-world deployment in dynamic moderation environments, where adaptability and coverage are critical.

\subsection{Ablations}
Our ablation study (Table \ref{tab:ablation}) isolates the impact of each component of our methodology. 

First, removing our synthetic dataset causes a catastrophic drop in performance on our domain-specific benchmark, RobloxGuard-Eval from 79.6\% F1 down to 20.3\%. Performance also drops significantly on OAI Mod from 70.3\% to 49.2\% and WildGuard Response from 80.6\% to 68.3\%. This confirms that public datasets alone are insufficient to cover the nuanced, fine-grained policies prevalent in real-world applications and highlights the critical value of our synthetic data pipeline.

Second, removing CoT rationales leads to performance degradation on several complex benchmarks requiring deeper reasoning, notably Aegis 2.0 Response (a 4.4\% drop) and Harmbench (a 3.9\% drop). Interestingly, performance slightly improved on SafeRLHF and RobloxGuard-Eval. We hypothesize this occurs because these benchmarks may contain a higher proportion of straightforward violations where explicit CoT reasoning provides less benefit compared to simpler pattern matching. Overall, the results suggest CoT is most beneficial for nuanced, multi-turn, or reasoning-intensive safety evaluations.

Finally, removing input inversion most significantly impacts performance on XSTest (a 3.0\% drop) and WildGuard Response (a 2.6\% drop). Since XSTest evaluates model brittleness and over-refusals, and WildGuard uses adversarial formats, this confirms that input inversion is key to improving model robustness and flexibility against diverse instruction formats.

\section{Conclusion}
We present Roblox Guard 1.0, a binary classification system operating at both prompt and response levels to ensure LLM safety. Our model exhibits strong generalization to unseen taxonomies and achieves state-of-the-art or comparable performance across multiple benchmarks. Central to our approach is a synthetic data pipeline augmented with CoT rationales, which enriches contextual grounding during training. Additionally, we introduce RobloxGuard-Eval, a comprehensive new evaluation dataset aligned with Roblox’s rigorous safety taxonomy, annotated by internal policy experts to ensure high-fidelity labels that capture nuanced policy distinctions. As existing safety benchmarks approach saturation, RobloxGuard-Eval provides a fresh, challenging benchmark designed to rigorously evaluate taxonomy-awareness and out-of-domain generalization in safety-critical contexts. We encourage the community to adopt RobloxGuard-Eval as a standard for future research in LLM safety and moderation.


\bibliography{aaai2026}


\end{document}